%% file: main.tex
\documentclass[10pt,twocolumn,letterpaper]{article}

\usepackage[pagenumbers]{cvpr} 

\usepackage[utf8]{inputenc} 
\usepackage{CJKutf8} 
\usepackage{mathrsfs}
\usepackage{amsmath}
\usepackage{bm}
\usepackage{multirow}
\usepackage{bbding}

\input{preamble}

\definecolor{cvprblue}{rgb}{0.21,0.49,0.74}
\usepackage[pagebackref,breaklinks,colorlinks,citecolor=cvprblue]{hyperref}

\def\paperID{1526} 
\def\confName{CVPR}
\def\confYear{2024}

\usepackage[capitalize]{cleveref}
\crefname{section}{Sec.}{Secs.}
\Crefname{section}{Section}{Sections}
\Crefname{table}{Table}{Tables}
\crefname{table}{Tab.}{Tabs.}
\Crefname{figure}{Figure}{Figures}
\crefname{figure}{Fig.}{Figs.}
\Crefname{equation}{Equation}{Equations}
\crefname{equation}{Eq.}{Eqs.}

\newcommand{\algname}{Upscale-A-Video~}
\newcommand{\eat}[1]{}

\newcommand{\blue}[1]{{\textcolor{blue}{#1}}}

\usepackage{hyperref}
\usepackage{url}
\usepackage{footnote}
\hypersetup{
colorlinks,
linkcolor={red},
urlcolor={purple}
}

\title{Upscale-A-Video: Temporal-Consistent Diffusion Model \\for Real-World Video Super-Resolution}

\author{Shangchen Zhou$^{*}$ \quad Peiqing Yang$^{*}$ \quad Jianyi Wang \quad Yihang Luo \quad Chen Change Loy\\
S-Lab, Nanyang Technological University\\
{\tt\small \url{https://shangchenzhou.com/projects/upscale-a-video}}
}

\begin{document}
\begin{CJK}{UTF8}{gbsn}

\twocolumn[{%
\renewcommand\twocolumn[1][]{#1}%
\maketitle
\vspace{-8mm}
\begin{center}
    \centering
    \includegraphics[width=\linewidth]{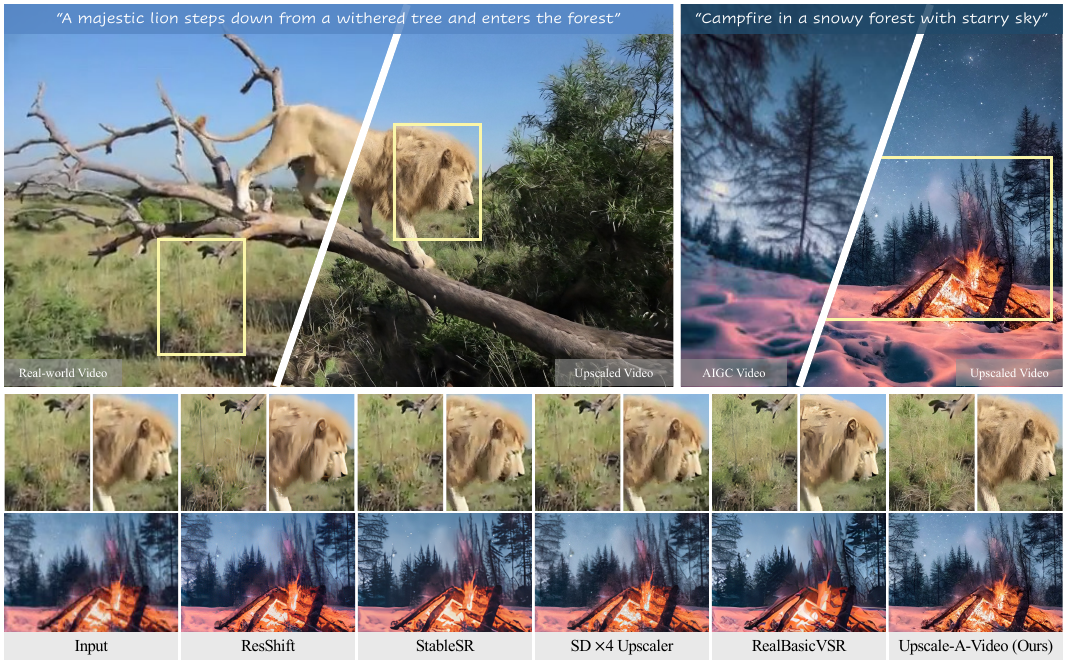}
    \vspace{-6mm}
    \captionof{figure}{
    Video super-resolution comparisons on both real-world and AI-generated videos. Our proposed \algname showcases excellent upscaling capabilities. By using appropriate text prompts, it achieves impressive results characterized by finer details and heightened visual realism.
    \textbf{(Zoom-in for best view)}
    } \vspace{1mm}
    \label{fig:teaser}
\end{center}%
}]
\def\thefootnote{*}\footnotetext{Equal contribution.}

%
\input{0_abstract}    
\input{1_introduction}
\input{2_relatedwork}
\input{3_method}
\input{4_experiment}
\input{5_conclusion}

\clearpage

\noindent{\bf Acknowledgement.} This study is supported under the RIE2020 Industry Alignment Fund Industry Collaboration Projects (IAF-ICP) Funding Initiative, as well as cash and in-kind contribution from the industry partner(s). The project is also supported by Shanghai AI Laboratory.

{
    \small
    \bibliographystyle{ieeenat_fullname}
    \bibliography{main}
}

\input{X_suppl}
\end{CJK}
\end{document}

%% file: preamble.tex
\usepackage[dvipsnames]{xcolor}
\newcommand{\red}[1]{{\color{red}#1}}


%% file: 0_abstract.tex

\begin{abstract}
Text-based diffusion models have exhibited remarkable success in generation and editing, showing great promise for enhancing visual content with their generative prior. However, applying these models to video super-resolution remains challenging due to the high demands for output fidelity and temporal consistency, which is complicated by the inherent randomness in diffusion models.  Our study introduces \textbf{\text{Upscale-A-Video}}, a text-guided latent diffusion framework for video upscaling. This framework ensures temporal coherence through two key mechanisms: locally, it integrates temporal layers into U-Net and VAE-Decoder, maintaining consistency within short sequences; globally, without training, a flow-guided recurrent latent propagation module is introduced to enhance overall video stability by propagating and fusing latent across the entire sequences. Thanks to the diffusion paradigm, our model also offers greater flexibility by allowing text prompts to guide texture creation and adjustable noise levels to balance restoration and generation, enabling a trade-off between fidelity and quality. Extensive experiments show that \textbf{\text{Upscale-A-Video}} surpasses existing methods in both synthetic and real-world benchmarks, as well as in AI-generated videos, showcasing impressive visual realism and temporal consistency.
\end{abstract}

%% file: 1_introduction.tex

\section{Introduction}
\label{sec:intro}
Video super-resolution (VSR) in real-world scenarios is a challenging task that aims at enhancing the quality of low-quality videos to produce high-quality results. Unlike previous works that mainly focus on either synthetic degradations~\cite{wang2019edvr, chan2021basicvsr, chan2022basicvsr++} or specific camera-related degradations~\cite{yang2021real}, this task is more demanding due to the need for addressing complex and unknown degradations commonly found in low-quality videos, such as downsampling, noise, blur, flickering, and video compression. In addition, maintaining the visual fidelity and ensuring temporal coherence are crucial for the perceptual quality of the output video. Although recent convolutional neural network (CNN)-based networks~\cite{chan2022investigating, xie2023mitigating} have shown success in mitigating many forms of degradation, they still fall short in producing realistic textures and details due to their limited generative capabilities, often resulting in over-smoothing, which can be observed in \text{RealBasicVSR} results shown in Fig.~\ref{fig:teaser}.
Diffusion models~\cite{ho2020denoising} have exhibited impressive proficiency in generating high-quality images~\cite{saharia2022photorealistic, rombach2022high, ramesh2022hierarchical, zhang2023adding} and videos~\cite{ho2022imagen, blattmann2023align, zhou2022magicvideo, ge2023preserve, chen2023videocrafter1}. Harnessing their generative potential, diffusion-based models have been proposed for image restoration, including both training from scratch~\cite{rombach2022high, sahak2023denoising, yue2023resshift} and fine-tuning from pretrained Stable Diffusion models~\cite{wang2023exploiting, lin2023diffbir}. These methods have effectively mitigated the over-smoothing issue often observed in CNN-based models, yielding results with more realistic fine-grained details. However, adapting these diffusion priors to VSR remains a non-trivial challenge. This difficulty stems from the inherent randomness in diffusion sampling, which inevitably introduces unexpected temporal discontinuities in resulting videos. This issue is more pronounced in latent diffusion, where the VAE decoder further introduces flickering in low-level texture details. 
Recent efforts have been made to adapt image diffusion models for video tasks by introducing strategies for temporal consistency, which include: 1) Fine-tuning video models with temporal layers such as 3D convolution~\cite{blattmann2023align, wang2023videocomposer} and temporal attention~\cite{blattmann2023align, zhou2022magicvideo, chen2023videocrafter1, wang2023videocomposer}; 2) Employing zero-shot mechanisms like cross-frame attention~\cite{wu2023tune, yang2023rerender} and flow-guided attention~\cite{geyer2023tokenflow, cong2023flatten} in the pretrained models. Although these solutions significantly improve video stability, there remain two primary issues: i) The current methods, operating in the U-Net feature or latent space, struggle to maintain low-level consistency, with issues like texture flickering still present. ii) The existing temporal layers and attention mechanisms can only impose constraints on short, local input sequences, limiting their capacity to ensure global temporal consistency in longer videos.
To tackle these issues, we adopt a local-global strategy for maintaining temporal consistency in video reconstruction, focusing on both fine-grained textures and overall consistency. On a local video clip, we explore finetuning a pretrained image $\times 4$ upscaling model~\cite{sdupscaler} with additional temporal layers on the video data. Specifically, within a latent diffusion framework, we first finetune the U-Net with integrated 3D convolutions and temporal attention layers, and then tune the VAE-Decoder with video-conditioned inputs and 3D convolutions. The former significantly achieves structure stability in local sequences, while the latter further improves low-level consistency, reducing texture flickering. On a more global scale, we introduce a novel, training-free flow-guided recurrent latent propagation module. Spanning short video segments, it bidirectionally conducts frame-by-frame propagation and latent fusion during inference. Leveraging latent fusion in a long-term sequence, this module encourages overall stability for long videos.
We further investigate the generative potential of diffusion models in the VSR task. Following the text-guided fashion, our model can leverage text prompts as an optional condition to steer the model towards producing more realistic and high-quality details, as illustrated in Fig.~\ref{fig:teaser}. Furthermore, we enhance the robustness of our model against heavy or unseen degradation by injecting noise into input to dilute its degradation. By adjusting the level of added noise, we can modulate the balance between restoring and generating within the diffusion model. Lower noise levels prioritize the model's restoration capabilities, while higher levels encourage generation of more refined details, thus achieving a trade-off between fidelity and quality of the output.
To summarize, the main contribution of our study is a practical and robust approach to real-world VSR. In particular, we conduct an exhaustive exploration of integrating a local-global temporal strategy within a latent diffusion framework, enhancing temporal coherence and generation quality. We further examine ways to improve versatility by allowing text prompts to guide texture creation, and offer control over noise levels to balance restoration and generation, thereby achieving a trade-off between fidelity and quality. Thanks to our design and pretrained generative prior, our model achieves state-of-the-art performance on existing benchmarks, showing remarkable visual realism and temporal consistency.

%% file: 2_relatedwork.tex

\section{Related Work}
\label{sec:related_work}

\noindent {\bf Video Super-Resolution.}
VSR aims to restore a sequence of high-resolution (HR) video frames from its degraded low-resolution (LR) counterparts.
Most existing approaches~\cite{liang2022rvrt,liang2022vrt,cao2021video,chan2021basicvsr,chan2022basicvsr++,isobe2020video1,isobe2020video,isobe2020revisiting,jo2018deep,wang2019edvr,xue2019video} assume a pre-defined degradation process \cite{liu2013bayesian,nah2019ntire,xue2019video,yi2019progressive}, and their performance deteriorates significantly in real-world scenarios due to the limited generalizability.
To tackle real-world VSR, recent works go beyond the traditional paradigms by assuming inputs with unknown degradations.
Due to the lack of real-world paired data for training, Yang \etal \cite{yang2021real} propose to collect HR-LR data pairs with iPhone cameras.
While the VSR model trained on such data can be effective to videos captured by similar mobile cameras, its generalization capability to other devices is in doubt.
Rather than relying on labor-intensive pair data collection, recent studies~\cite{chan2022investigating,xie2023mitigating} have shifted towards employing diverse degradations for data augmentation during training, demonstrating better performance in handling real-world cases.
Nonetheless, it is still challenging for existing CNN-based approaches~\cite{chan2022investigating,xie2023mitigating} to generate photo-realistic textures due to the absence of generative prior.
In this study, we set our sights on exploiting robust and extensive generative prior encapsulated in a pretrained image diffusion model, \ie, Stable Diffusion (SD) $\times$4 upscaler \cite{sdupscaler}.
By integrating this strong diffusion prior, our approach circumvents the need for exhaustive training from scratch and exhibits improved performance in producing detailed textures.

\noindent {\bf Diffusion Models for Video Tasks.}
Following the success of text-to-image diffusion models~\cite{choi2021ilvr,avrahami2022blended,hertz2022prompt,gu2022vector,mou2023t2i,zhang2023adding,gal2023designing}, recent studies have ventured into the applications of video diffusion models~\cite{ho2022video,ho2022imagen,luo2023videofusion,lu2023vdt,hu2023lamd,mei2023vidm,esser2023structure}.
Instead of training from scratch, some methods~\cite{qi2023fatezero,wu2023tune,yang2023rerender,tokenflow2023,cong2023flatten} focus on video generation using off-the-shelf image diffusion models~\cite{rombach2022high,zhang2023adding} in a zero-shot manner.
To keep temporal consistency, cross attention~\cite{vaswani2017attention} between neighboring frames and optical flow~\cite{teed2020raft} warping are popular solutions adopted in these methods.
While being efficient, these methods generally suffer from limited generalizability and the selection of hyperparameters can be tricky.
Most recently, Blattmann \etal~\cite{blattmann2023align} propose to extend pretrained image diffusion models to the video domain by introducing an additional temporal dimension and fine-tuning the temporal layers only to speed up the training process.
Subsequent works~\cite{guo2023animatediff,xing2023simda} following such paradigm are capable of generating impressive video sequences.
Inspired by these works, we employ a pretrained image $\times 4$ upscaling model~\cite{sdupscaler} as generative prior and propose a novel local-global temporal strategy, resulting in temporally coherent outputs with faithful details.


%
\noindent {\bf Diffusion Models for Restoration Tasks.}
With notable advancements in diffusion models~\cite{sohl2015deep,ho2020denoising,2021Score,nichol2022glide}, numerous diffusion-based works have been proposed for image restoration.
A straightforward way is to train a diffusion model conditioned on LR images from scratch~\cite{yue2023resshift,xia2023diffir,rombach2022high,saharia2022image,sahak2023denoising}, which, however, demands significant computational resources for training.
To avoid such heavy training costs, another prevalent approach~\cite{yang2023pgdiff,choi2021ilvr,wang2022zero,chungimproving,songpseudoinverse,kawar2022denoising,wang2023ddnm} is to incorporate constraints into the reverse diffusion process of a pretrained diffusion model.
Though efficient, designing these constraints relies on pre-defined image degradation processes or pretrained SR models as a priori, leading to limited generalizability and inferior performance.
Recent works~\cite{wang2023exploiting,yang2023pixel} go further by fine-tuning directly on a frozen pretrained diffusion model with several additional trainable layers, demonstrating impressive performance.
Inspired by these recent advances, we focus on exploiting the effective diffusion prior for real-world VSR, which remains under-explored and challenging due to the temporal discontinuities caused by inherent randomness during diffusion sampling.
With the proposed local-global temporal strategy, our approach is capable of generating temporally coherent results for real-world VSR.

%% file: 3_method.tex

\section{Methodology}
\label{sec:method}
\begin{figure*}[th]
\begin{center}
    \includegraphics[width=.99\linewidth]{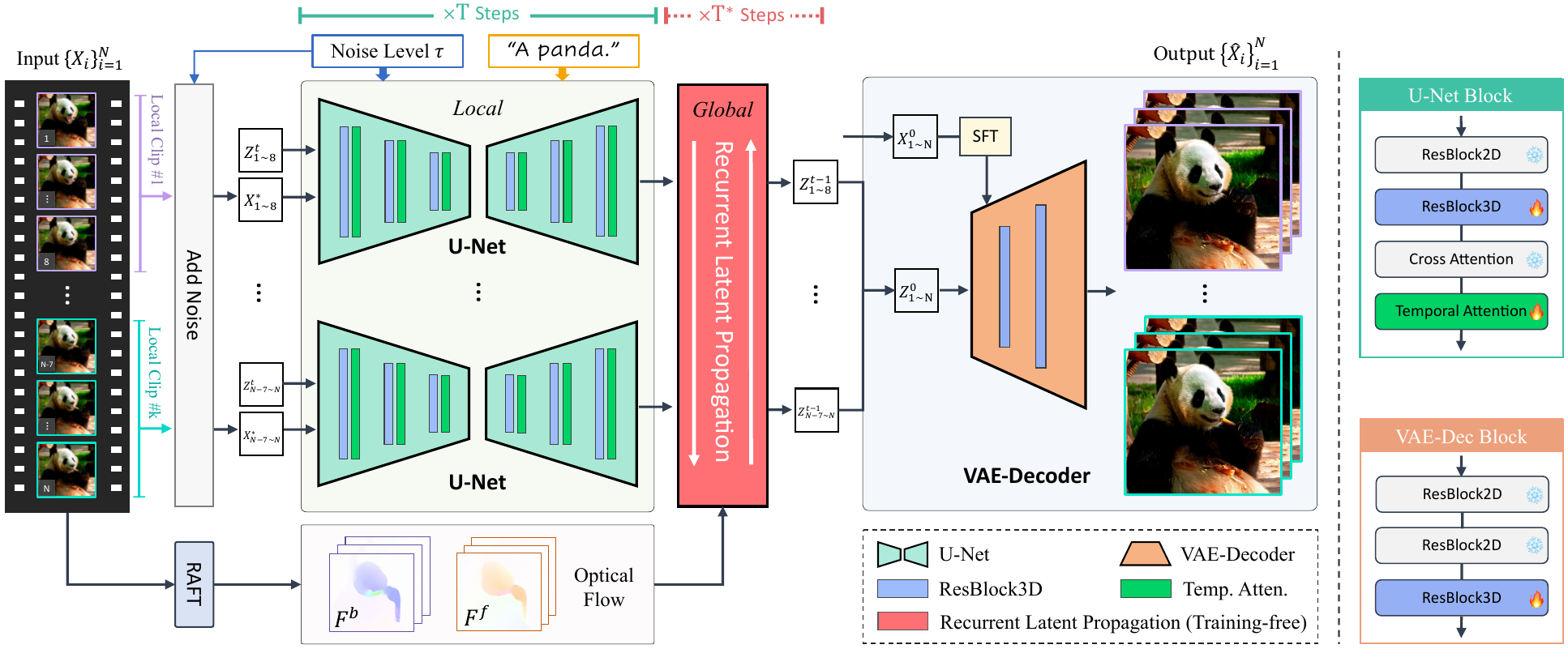}
    \vspace{-1mm}
    \caption{
    An overview of Upscale-A-Video.
    Upscale-A-Video processes long videos using both local and global strategies to maintain temporal coherence. It divides the video into segments and processes them using a U-Net with temporal layers for intra-segment consistency.
    During user-specified diffusion steps for global refinement, a recurrent latent propagation module is used to enhance inter-segment consistency.
    Finally, a finetuned VAE-Decoder reduces remaining flickering artifacts for low-level consistency.
    Our model also allows users to guide texture creation with text prompts and adjust noise levels to balance the effect of restoration and generation.
    }
    \vspace{-6mm}
\label{fig:overview}
\end{center}
\end{figure*}

Our objective is to develop a text-guided diffusion framework tailored for real-world VSR. The diffusion denoising process, characterized by its inherent stochastic nature, poses significant challenges when applied to video tasks. These challenges include temporal instability and the emergence of flickering artifacts, which are particularly prominent in VSR tasks involving lengthy video sequences. The complexity of these tasks lies not just in achieving temporal consistency within localized segments, but also in preserving coherence throughout the entire video.
As illustrated in Fig.~\ref{fig:overview}, our framework incorporates both local and global modules in the latent diffusion model (LDM) to preserve temporal consistency both within and across video segments.
Within each diffusion time step ($t=1,2\dots T$), the video is split into segments and processed with a U-Net that includes temporal layers to ensure consistency within each segment. If the current time step falls within the user-specified global refinement steps ($T^*$), a recurrent latent propagation module is employed to improve consistency across segments during inference. Lastly, a finetuned VAE-Decoder is used to reduce remaining flickering artifacts.
Thanks to the diffusion paradigm, our model exhibits remarkable versatility: 
1) Input text prompts can enhance video quality further with improved realism and details. 
2) User-specified noise levels provide a control over the trade-off between quality and fidelity.
\subsection{Preliminary: Diffusion Models}
\noindent {\bf Pretrained Stable Diffusion Image $\times 4$ Upscaler.}
Our \text{Upscale-A-Video} is built upon the pretrained text-guided SD $\times 4$ Upscaler~\cite{sdupscaler}. It employs the LDM framework~\cite{rombach2022high}, which uses an autoencoder to transform an image into a latent with an encoder $\mathcal{E}$ ($\times$4 downsampling), and reconstructs it with a decoder $\mathcal{D}$. Conditioned on the low-resolution images $\text{x}$, it learns to generate the high-quality counterparts via iterative denoising in the latent space. 
To achieve this, given latent samples $\text{z} \sim p_{data}$, diffused latents $\text{z}_t = \alpha_t \text{z} + \sigma_t \bm{\epsilon}$ is generated by introducing Gaussian noise to the latents $\text{z}$ at each diffusion step $t$; where $\bm{\epsilon} \sim \mathcal{N}(\bm{0}, \bm{I})$, $\alpha_t$ and $\sigma_t$ define a noise schedule. 
To enhance the ability of our model to generate fine details, it also applies random noise to input images, \ie, the diffused images $\text{x}_{\tau} = \alpha_{\tau} \text{x} + \sigma_{\tau} \bm{\epsilon}$, where $\tau$ is the noise level that corresponds to the early steps in noise schedule.
Adopting $\text{v}$-prediction perameterization~\cite{salimans2022progressive}, the U-Net denoiser $f_\theta$ is trained to make predictions of $\text{v}_t \equiv \alpha_t \bm{\epsilon} - \sigma_t \text{x}$. The optimization objective for LDM is as follows:
\begin{equation}
\mathbb{E}_{\text{z},\text{x},\text{c},t,\bm{\epsilon}}\left[ \| \text{v} -f_{\theta}(\text{z}_t, \text{x}_{\tau}; \text{c},t)\|_2^2\right], 
\label{eq:denoising_objective}
\end{equation}
where $\text{c}$ serves as an optional set of conditions, including text prompts and noise levels of the diffused image. During inference, the model has the flexibility to involve different text prompts and noise levels to diffusion sampling of $x_0$, and finally decode it to produce the $\times$4 upscaled images.
\noindent {\bf Inflated 2D Convolution.}
When adapting pretrained 2D diffusion models to video tasks, it is common to inflate their 2D convolutions into the 3D convolutions~\cite{blattmann2023align, ge2023preserve, wu2023tune, yang2023rerender}. This modification allows the diffusion models to smoothly integrate new temporal layers, enabling them to capture and encode temporal information within the pretrained model.
In this study, we explore a new avenue to create a text-guided video diffusion model for VSR tasks, starting with the pretrained SD $\times 4$ Upscaler~\cite{sdupscaler}. To process video data, we first modify its network structure by inflating the 2D convolutions into 3D convolutions, and then initialize our network with this upscaler to inherit its enhancement capabilities. Our goal is to transfer the knowledge learned from image upscaling to video enhancement, enabling more efficient training. In the following, we we will describe our local-global strategy within the LDM framework to achieve temporal coherence while harnessing the capability of diffusion for VSR tasks.
\subsection{Local Consistency within Video Segments}
To apply the pretrained text-to-image SD model to video-related tasks, existing video diffusion models typically employ techniques such as 3D convolutions~\cite{blattmann2023align, wang2023videocomposer}, temporal attention~\cite{blattmann2023align, wang2023videocomposer, ge2023preserve, wang2023lavie}sx, and cross-frame attention~\cite{wu2023tune} to ensure temporal consistency.
\noindent {\bf Finetuning Temporal U-Net.}
Following the existing methods~\cite{blattmann2023align, wang2023videocomposer}, we introduce additional temporal layers into the pretrained image model and learn the local consistency constraint within video segments. As illustrated in Fig.~\ref{fig:overview}, in the modified temporal U-Net, we opt for temporal attention and 3D residual blocks based on 3D convolutions to serve as our temporal layers, and insert them within the pretrained spatial layers.
The temporal attention layer performs self-attention along the temporal dimension and focuses on all local frames. Additionally, we add Rotary Position Embedding (RoPE)~\cite{llama} into the temporal layers to provide the model with positional information for time.
Our temporal layers are trained using the same noise schedule as the pretrained image model. Importantly, we keep the pretrained spatial layers fixed during training and only optimize the inserted temporal layers using Eq.~\eqref{eq:denoising_objective}. An essential benefit of this strategy is that it allows us to leverage the pretrained spatial layers that were learned from a huge, high-quality image dataset. This enables us to concentrate our training efforts on refining the temporal layers.
\noindent {\bf Finetuning Temporal VAE-Decoder.}
Even after finetuning the U-Net on video data, the VAE-Decoder within the LDM framework, which is trained on images only, still tends to produce flickering artifacts when decoding a latent sequence. To mitigate this issue, we introduce additional temporal 3D residual blocks into the VAE-Decoder to enhance low-level consistency. 
Furthermore, the diffusion denoising process in the U-Net often introduces color shifts, a problem also encountered by other diffusion-based restoration networks~\cite{wang2023exploiting, yue2023resshift}.  To tackle this, we condition the input videos with a Spatial Feature Transform (SFT) layer~\cite{wang2018recovering}, which employs the inputs to transform the features of the first layer of the VAE-Decoder. This allows the input videos to provide low-frequency information, such as color, to strengthen the color fidelity of the output results.
Similar to the training of the Temporal U-Net, we keep the pretrained spatial layers unchanged and only train the newly added temporal layers. These temporal layers are trained on video data using a hybrid loss consisting of L1 loss, LPIPS perceptual loss~\cite{zhang2018unreasonable}, and an adversarial loss employing a temporal PatchGAN~\cite{zhou2023propainter} discriminator.
As demonstrated by the ablation study in Fig.~\ref{fig:temporal} and Table~\ref{tab:ablation}, this step is crucial for achieving favorable results.
\subsection{Global Consistency cross Video Segments}
\label{sec:prop}
\begin{figure}[t]
\begin{center}
    \includegraphics[width=.99\linewidth]{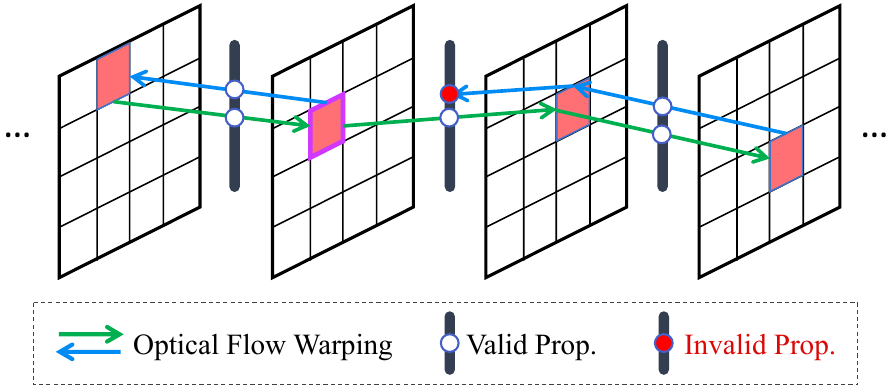}
    \vspace{0mm}
    \caption{
    An illustration of flow-guided recurrent latent propagation.
    Without requiring any learning, this module can achieve coherence across video segments via long-term latent propagation and aggregation. It relies on optical flow validity determined by forward-backward consistency error~\cite{meister2018unflow}. Only latent positions with low consistency errors will be propagated, while those with high errors, marked with a red dot, are not.
    }
    \vspace{-5mm}
\label{fig:prop}
\end{center}
\end{figure}

The trained temporal layers within the LDM are limited to processing local sequences (\eg, eight frames in our U-Net setting), making it impossible to enforce global consistency constraints across video segments. 
Prior studies~\cite{chan2022basicvsr++, chan2022investigating, zhou2023propainter, zhou2019spatio} have already showcased the benefits of flow-guided long-term propagation in enhancing temporal consistency for video restoration tasks. 
However, their performance gains are often observed when dealing with long video sequences that provide long-term information. Unfortunately, these methods are not well-suited for diffusion models due to memory constraints, which typically restrict them to processing short video clips.
\noindent {\bf Training-Free Recurrent Latent Propagation.}
We introduce a \textit{training-free} flow-guided \textit{recurrent} propagation module within the latent space. This module ensures global temporal coherence for long input videos, 
involving \textit{bidirectional} propagation in the forward and backward directions. Here, we elaborate on the forward propagation, and backward propagation follows the same process.
Given an input low-resolution video, we first adopt RAFT~\cite{teed2020raft} to estimate optical flow, with its resolution exactly matching the latent resolution, thus no need for resizing. We then check the validity of the estimated flow by evaluating forward-backward consistency error~\cite{meister2018unflow}: 
\begin{equation}
{E}_{i-1\rightarrow i}\big(p\big) = \Big\| {f}_{i-1\rightarrow i}\big(p\big) +{f}_{i\rightarrow i-1}\big(p+{f}_{i-1\rightarrow i}(p)\big)\Big\|_2^2,
\label{eq:flow_check}
\end{equation}
where $p$ denotes the position of the last frame latent, ${f}_{i-1\rightarrow i}$ and ${f}_{i\rightarrow i-1}$ are forward and backward flow respectively. As shown in Fig.~\ref{fig:prop}, only latents with small consistency error will be propagated, which can be thought of as an occlusion mask, $M: E_{i-1\rightarrow i}(p) < \delta $, where $\delta$ is a threshold. 
Let $\text{z}_i$ be the latent feature for the $i$-th frame at diffusion step $t$, we update the predicted $\hat{\text{z}}_0$ as $\widetilde{\hat{\text{z}}_0}$
The process of \textit{recurrent} latent propagation and aggregation is expressed as:
%
\begin{align}
\widetilde{\hat{\text{z}}_0^i} &= \big[\mathcal{W}(\widetilde{\hat{\text{z}}_0^{i-1}}, {f}_{i\rightarrow i-1}) * \beta + \hat{\text{z}}_0^i *  (1-\beta)\big]* M \nonumber \\
&\quad + \hat{\text{z}}_0^i * (1-M), \label{eq:latent_prop}
\end{align}
where $\mathcal{W}(\cdot)$ denotes warping operation (with \texttt{nearest} mode), and $\beta \in [0, 1]$ serves as a fusion weight for aggregating the warped latent from the previous $(i-1)^{th}$ frame to the current $i^{th}$ frame, where smaller values tend to preserve current information, while larger values favor the propagated information. We set $\beta$ to 0.5 by default.

It is not necessary to apply this module at each diffusion step during the inference process. Instead, we can choose $T^*$ steps for latent propagation and aggregation. When dealing with minor video jitter, one can opt to integrate this module early in the diffusion denoising process, while for severe video jitter, such as AIGC videos, it is preferable to execute this module later in the denoising process.

\begin{table*}[th]
    \begin{center}
    \caption{
        Quantitative comparisons on different VSR benchmarks from diverse sources, \ie, synthetic (SPMCS, UDM10, REDS30, YouHQ40), real (VideoLQ), and AIGC (AIGC30) data. The best and second performances are marked in \red{\underline{red}} and \blue{{blue}}, respectively.
        $E^*_{warp}$ denotes $E_{warp}~(\times 10^{-3})$.
        }
    \label{tab:comparison}
    \vspace{-2mm}
    \renewcommand{\arraystretch}{1.15}
    \renewcommand{\tabcolsep}{1.8mm}
    \scalebox{0.71}{
    \begin{tabular}{l|c|c|c|c|c||c|c|c|c}
    \hline
    Datasets & Metrics & Real-ESRGAN~\cite{wang2021realesrgan} & SD $\times$4 Upscaler~\cite{sdupscaler} & ResShift~\cite{yue2023resshift} & StableSR~\cite{wang2023exploiting} & RealVSR~\cite{yang2021real}  & DBVSR~\cite{pan2021deep} & RealBasicVSR~\cite{chan2022investigating} & Ours \\
    \hline
    \hline
    \multirow{4}{*}{SPMCS} & PSNR $\uparrow$ &	22.89	&23.19	&23.27&	22.71	&23.88&	24.28	&\blue{24.51}	&\red{\underline{25.32}} \\
         & SSIM $\uparrow$ 	&0.669	&0.631&	0.667&	0.657&	0.681&	\blue{0.726}&	0.717&	\red{\underline{0.741}} \\
         & LPIPS $\downarrow$ &	0.238&	0.304&	0.257&	0.231&	0.437&	0.302&	\red{\underline{0.198}}&	\blue{0.222} \\
         & $E^*_{warp} \downarrow$ &	1.364&	5.008&	4.942&	4.815&	\red{\underline{0.294}}&	1.360&	0.559&	\blue{0.367} \\
    \hline
    \multirow{4}{*}{UDM10} & PSNR $\uparrow$ 	&27.13	&28.07&	27.62	&26.45&	27.38&	\blue{29.60}&	29.11	&\red{\underline{30.79}} \\
         & SSIM $\uparrow$ 	&0.843	&0.811	&0.827&	0.825	&0.825&	\red{\underline{0.880}}&	0.876&	\blue{0.878} \\
         & LPIPS $\downarrow$ 	&0.190	&0.186&	0.222	&0.181	&0.278	&\blue{0.155}&	0.172	&\red{\underline{0.133}} \\
         & $E^*_{warp} \downarrow$ &	1.462	&1.710	&2.196	&2.797	&\blue{0.531}	&1.943	&0.602	&\red{\underline{0.446}} \\
    \hline
    \multirow{4}{*}{REDS30} & PSNR $\uparrow$ 	&22.40&	22.98&	23.00&	23.72&	23.05&	\blue{24.37}&	23.91&	\red{\underline{24.41}} \\
         & SSIM $\uparrow$ &	0.591	&0.572	&0.580	&\blue{0.635}	&0.603&	0.633&	\red{\underline{0.636}}&	0.631 \\
         & LPIPS $\downarrow$ &	\blue{0.303}&	0.399&	0.369&	0.352&	0.658	&0.588&	\red{\underline{0.249}}&	0.335 \\
         & $E^*_{warp} \downarrow$ &	3.658&	3.753	&4.131&	1.645&	\red{\underline{0.378}}&	9.659	&1.557	&\blue{1.278} \\
    \hline
    \multirow{4}{*}{YouHQ40} & PSNR $\uparrow$ 	&24.37&	19.71&	23.77&	24.53&	24.19&	\blue{25.37}	&24.09	&\red{\underline{25.83}} \\
         & SSIM $\uparrow$ &	0.710	&0.579	&0.654&	0.711&	0.695	&\blue{0.719}&	0.689	&\red{\underline{0.733}} \\
         & LPIPS $\downarrow$ 	&0.272&	0.442	&0.376&	\blue{0.271}&	0.484	&0.430&	0.306	&\red{\underline{0.268}} \\
         & $E^*_{warp} \downarrow$ &	1.856&	3.399	&4.426	&1.529&	\red{\underline{0.485}}	&1.149&	1.052	&\blue{0.737} \\
    \hline
    \hline
    \multirow{2}{*}{VideoLQ} & CLIP-IQA $\uparrow$&	0.360	&0.158&	\blue{0.430}&	0.344&	0.211&	0.274&	0.387	&	\red{\underline{0.530}}\\
         & MUSIQ $\uparrow$	&49.48	&26.21	&40.95	&44.23&	24.52	&29.15&	\blue{55.33}		&\red{\underline{57.99}} \\
         & DOVER $\uparrow$ 	&7.161&	2.884&	4.679&	6.783&	2.531	&3.628&	\blue{7.562}	&	\red{\underline{7.811}} \\
    \hline
    \multirow{2}{*}{AIGC30} & CLIP-IQA $\uparrow$ &	0.430	&0.329	&\blue{0.569}&	0.467&	0.276&	0.290	&0.565	&	\red{\underline{0.674}} \\
         & MUSIQ $\uparrow$ &	47.09	&35.30&	43.32	&44.93&	24.39&	27.22&	\red{\underline{58.87}}	&	\blue{57.66} \\
         & DOVER $\uparrow$ &	9.710&	5.646	&7.042	&9.668&	3.285 &	3.523	&\blue{10.68}	&	\red{\underline{11.67}} \\
    \hline
    \end{tabular}
    }
    \end{center}
    \vspace{-5mm}
\end{table*}

\subsection{Inference with Additional Conditions}
We can further adjust additional conditions of \textit{text prompts} and \textit{noise levels} in Upscale-A-Video to influence the diffusion denoising process. Text prompts can guide the generation of texture details, such as animal fur or oil painting strokes, as shown in Fig.~\ref{fig:text_prompt}. 
Besides, adjusting the noise level allows us to balance the model's restoration and generation abilities, with smaller values favoring restoration and larger values promoting the generation of more details (see comparisons in Fig.~\ref{fig:conditions}).
We also adopt Classifier-Free Guidance (CFG)~\cite{ho2022classifier} during inference, which can significantly enhance the impact of both text prompts and noise levels, helping produce high-quality videos with finer details.
%

%% file: 4_experiment.tex

\section{Experiments}
\label{sec:experiment}
\subsection{Datasets and Implementation}
\noindent {\bf Training Datasets.}
We train our Upscale-A-Video using the following datasets: 1) The subset of WebVid10M ~\cite{webvid} contains around 335K video-text pairs with each resolution around $336\times 596$, which is commonly used in training video diffusion models~\cite{blattmann2023align,ge2023preserve,zhang2023show,chen2023videocrafter1}; 2) YouHQ dataset. Due to the lack of high-quality video data for training, we additionally collect a large-scale high-definition ($1080\times 1920$) dataset from YouTube, containing around 37K video clips with diverse scenarios, \ie, street view, landscape, animal, human face, static object, underwater, and nighttime scene.
Training on such high-quality data further enhances the generation ability of our model for real-world VSR. 
Following the degradation pipeline of RealBasicVSR~\cite{chan2022investigating}, we generate the LQ-HQ video pairs for training.
\noindent {\bf Testing Datasets.}
For synthetic testing datasets, we construct four synthetic datasets (\ie, SPMCS~\cite{PFNL}, UDM10~\cite{tao2017spmc}, REDS30~\cite{nah2019ntire}, and YouHQ40), which follow the same degradation pipeline in training to generate corresponding LQ videos. We split the YouHQ40 test set from the proposed YouHQ dataset, containing 40 videos.
Additionally, we evaluate the models on a real-world dataset (\ie, VideoLQ~\cite{chan2022investigating}) and an AIGC dataset (named AIGC30) that collects 30 AI-generated videos by popular text-to-video generation models~\cite{blattmann2023align,wang2023lavie,luo2023videofusion,ge2023preserve,zhang2023show,wang2022zero,ho2022imagen,pikalab,zeroscope}.

\noindent {\bf Training Details.}
Our Upscale-A-Video is trained on 32 NVIDIA A100-80G GPUs with a batch size of 384.
The training data is cropped to $80 \times 80$ with a length of $8$. The learning rate is set to $1 \times 10^{-4}$ using the Adam \citep{kingma2014adam} optimizer.
We first train the U-Net model on both WebVid10M~\cite{webvid} and YouHQ for 70K iterations.
Then we train another 10K iterations on YouHQ only. 
Since there are no text prompts for YouHQ, we use the null prompts during training.
In this way, our model can handle VSR with either LQ inputs and prompts or LQ inputs only, leading to a more flexible use in practice.
As for VAE-Decoder fine-tuning, we follow StableSR~\cite{wang2023exploiting} to first generate 100K synthetic LQ-HQ video pairs on WebVid10M~\cite{webvid} and YouHQ, 
and the finetuned U-Net model is adopted to generate the corresponding latent codes for the LQ videos.

%
\noindent {\bf Evaluation Metrics.}
We adopt different metrics to evaluate both the frame quality as well as temporal coherency of the generated results.
For synthetic datasets with LQ-HQ pairs, we employ PSNR, SSIM, LPIPS~\cite{zhang2018unreasonable}, and the flow warping error~\cite{Lai-ECCV-2018} $E^*_{warp}$ for evaluation.
For real-world and AIGC test data, since no ground-truth videos are available, we conduct our evaluation on commonly used non-reference metrics, \ie, CLIP-IQA~\cite{wang2022exploring}, MUSIQ~\cite{ke2021musiq}, and DOVER~\cite{wu2023exploring}.
\begin{figure*}[t]
\begin{center}
    \includegraphics[width=.99\linewidth]{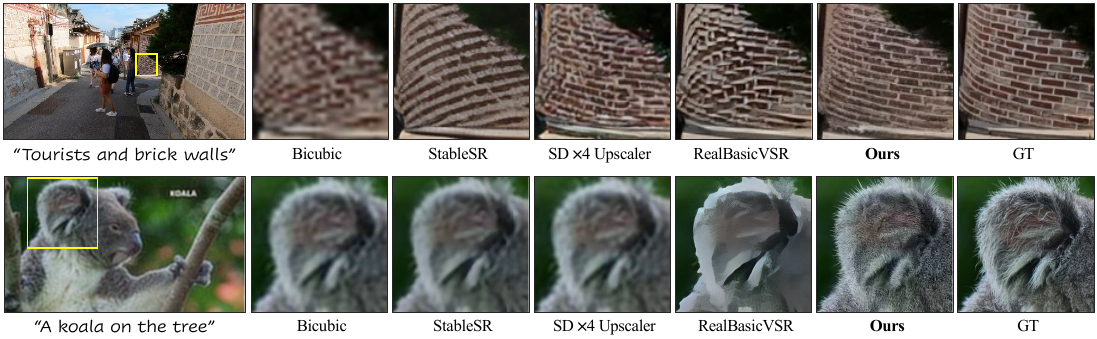}
    \vspace{-1mm}
    \caption{
    Qualitative comparisons on synthetic low-quality videos from REDS30~\cite{nah2019ntire} and YouHQ40 datasets. Among the tested methods, only our Upscale-A-Video can recover the accurate wall structure and produce detailed koala fur. 
    \textbf{(Zoom-in for best view)}
    }
    \vspace{-5mm}
\label{fig:qualitative_synthetic}
\end{center}
\end{figure*}
\begin{figure*}[t]
\begin{center}
    \includegraphics[width=\linewidth]{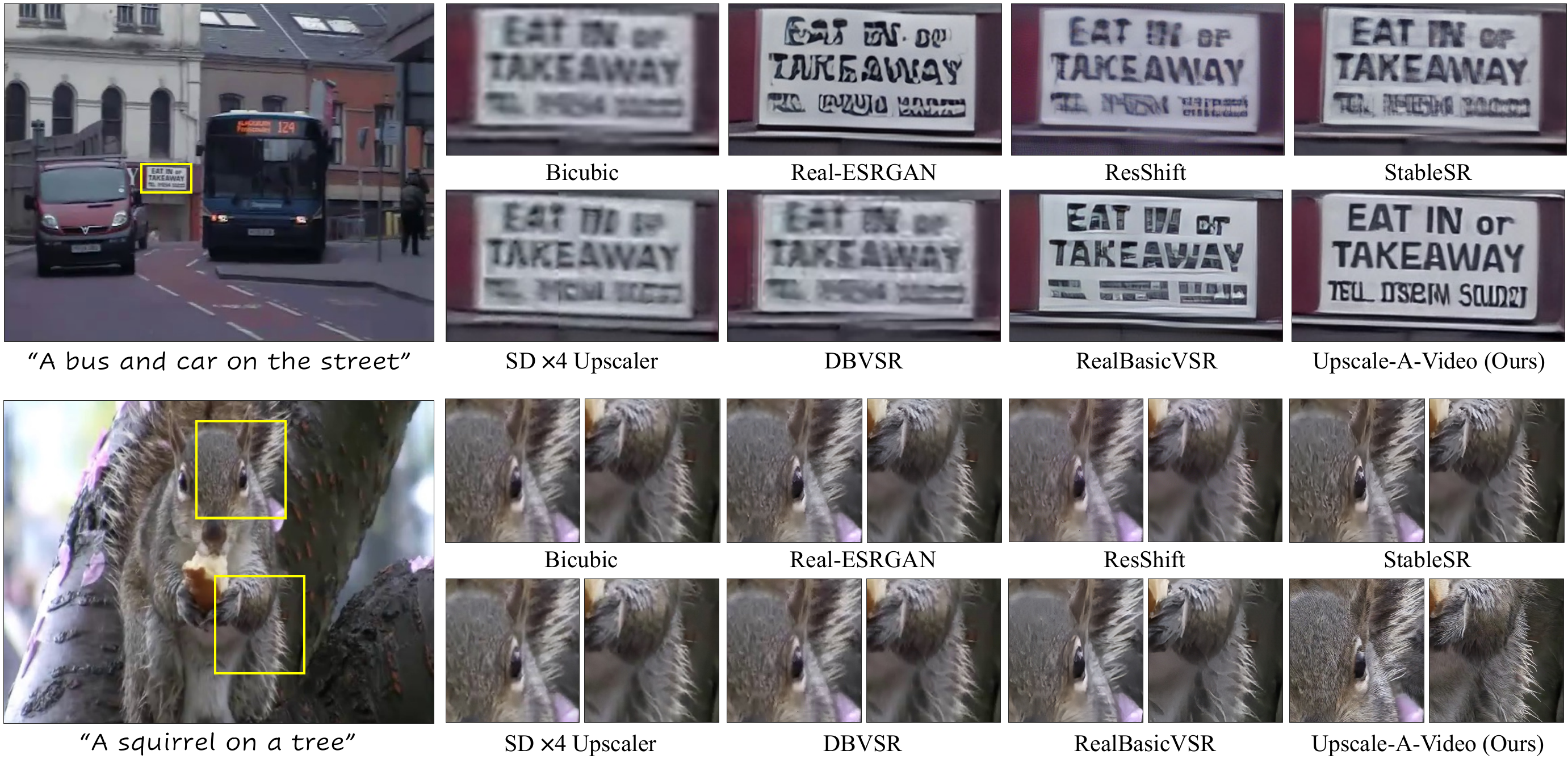}
    \vspace{-6mm}
    \caption{
    Qualitative comparisons on real-world test videos in VideoLQ~\cite{chan2022investigating} dataset. Our Upscale-A-Video effectively leverages the advantages of the diffusion paradigm in generating high-quality results.
    When compared to existing methods, it notably excels in its restoration capabilities, successfully recovering the billboard word ``EAT IN or TAKEAWAY''.
    In particular, when guided by text prompts, Upscale-A-Video showcases promising enhanced results with more details and heightened realism. 
    \textbf{(Zoom-in for best view)}
    }
    \vspace{-6mm}
\label{fig:qualitative_real}
\end{center}
\end{figure*}
\subsection{Comparisons}
To verify the effectiveness of our approach, we compare Upscale-A-Video with several state-of-the-art methods, including Real-ESRGAN~\cite{wang2021realesrgan}, SD $\times$4 Upscaler~\cite{sdupscaler}, ResShift~\cite{yue2023resshift}, StableSR~\cite{wang2023exploiting}, RealVSR~\cite{yang2021real}, DBVSR~\cite{pan2021deep}, and RealBasicVSR~\cite{chan2022investigating}.
%

%
\noindent {\bf Quantitative Evaluation.}
As shown in Table \ref{tab:comparison}, our approach achieves the highest PSNR across all four synthetic datasets, suggesting its outstanding reconstruction ability.
Moreover, our approach achieves the lowest LPIPS scores on both UDM10 and YouHQ40, indicating the high perceptual quality of our generated results.
In addition to the good performance on synthetic benchmarks, our approach further obtains the highest CLIP-IQA and DOVER scores on both the real dataset~\cite{chan2022investigating} and AIGC videos.
The superiority across datasets from various sources demonstrates the effectiveness of our approach.

\noindent {\bf Qualitative Evaluation.}
We present visual results on both synthetic~\cite{nah2019ntire,PFNL,tao2017spmc} and real-world~\cite{chan2022investigating} videos in Fig.~\ref{fig:qualitative_synthetic} and Fig.~\ref{fig:qualitative_real}, respectively.
It is observed that our Upscale-A-Video significantly outperforms existing CNN- and diffusion-based approaches in both artifact removal and details generation.
Specifically, Upscale-A-Video is capable of generating more natural details of the koala than other methods in Fig.~\ref{fig:qualitative_synthetic}, 
and successfully recovers the words ``EAT IN or TAKEAWAY'' on the billboard in the first example of Fig.~\ref{fig:qualitative_real}, while other methods generate blurry or distorted results.
Note that SD $\times$4 Upscaler shows less effectiveness for the real-world VSR since it does not consider mixed degradations in its training data.
%
%

%
\noindent {\bf Temporal Consistency.}
\begin{figure*}[t]
\begin{center}
    \vspace{-4mm}
    \includegraphics[width=.99\linewidth]{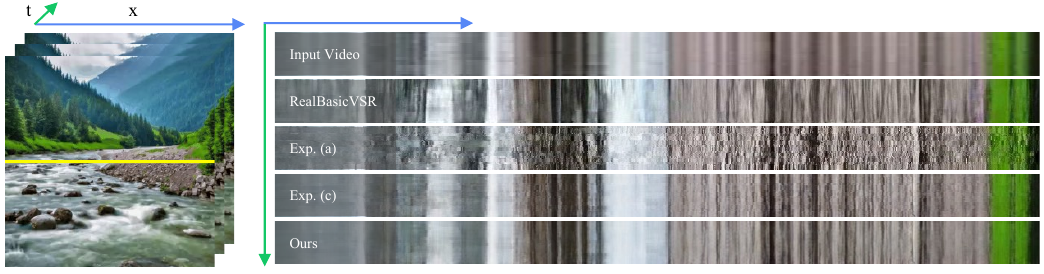}
    \vspace{-3mm}
    \caption{
    Comparison of temporal profile. We examine a row and track changes over time. The profile from existing methods and Exp. (a) and (c) (w/o latent propagation module) exhibit noise, suggesting the presence of flickering artifacts.  Even with the finetuned decoder, the profile of Exp. (c) still displays some discontinuities. Thanks to long-term latent propagation and aggregation, the profile of our Upscale-A-Video exhibits a more seamless and smoother transition.
    }
    \vspace{-6mm}
\label{fig:temporal}
\end{center}
\end{figure*}
Benefiting from our local-global temporal strategy, Upscale-A-Video achieves the best optical flow error score on UDM10 and the second-best scores on REDS30, SPMCS and YouHQ40, remarkably outperforming other diffusion-based approaches, and even beats the strong CNN-based VSR methods, \ie, RealBasicVSR and DBVSR.
We also visualize the temporal profile in Fig.~\ref{fig:temporal}.
It is observed that our approach achieves superior performance with a more seamless and smoother transition.
%
%
\subsection{Ablation Study}
\begin{table}[t]
\caption{Ablation study of finetuned VAE-Decoder and propagation module on YouHQ40.}
\centering
\vspace{-2mm}
\renewcommand{\arraystretch}{1.15}
\renewcommand{\tabcolsep}{2.05mm}
\resizebox{0.95\linewidth}{!} {
\begin{tabular}{c|cc|ccc}
\toprule
Exp. & ft-VAE-Dec. & Latent Prop. &    PSNR$\uparrow$ & SSIM$\uparrow$ & $E^*_{warp} \downarrow$\\ 
\midrule
(a)   &   & & 23.82& 0.6385 & 2.398\\
(b)  &  & \Checkmark  &25.47     &  0.7215  &  1.815 \\ 
(c) &\Checkmark&  &25.75 & {\bf 0.7328}    &  0.842    \\ 
(d) &\Checkmark & \Checkmark  &  {\bf 25.83}   &  0.7326 & {\bf 0.737}   \\ \bottomrule
\end{tabular}
}
\label{tab:ablation}
\vspace{-5mm}
\end{table}
\noindent {\bf Effectiveness of Finetuned VAE-Decoder.} 
We first investigate the significance of the fine-tuned VAE-Decoder.
As shown in Table \ref{tab:ablation}, replacing our finetuned VAE-Decoder with the original decoder leads to worse PSNR, SSIM, and $E^*_{warp}$.
Particularly, the increase of $E^*_{warp}$ from 0.737 to 1.815 indicates a significant deterioration of temporal coherency.
The comparisons in Fig.~\ref{fig:temporal} also suggest inferior temporal consistency without the finetuned VAE-Decoder.

\noindent {\bf Effectiveness of Propagation Module.} 
In addition to finetuning VAE-Decoder, our proposed flow-guided recurrent latent propagation module further enhances the stability of long videos.
As shown in Table \ref{tab:ablation}, adopting the propagation module can further reduce $E^*_{warp}$ error, effectively improving temporal consistency while maintaining high PSNR.
Similar phenomena can also be observed in the temporal profile in Fig.~\ref{fig:temporal}, showing a more seamless transition.

\noindent {\bf Text Prompt.} 
\begin{figure}[t]
\begin{center}
    \includegraphics[width=.99\linewidth]{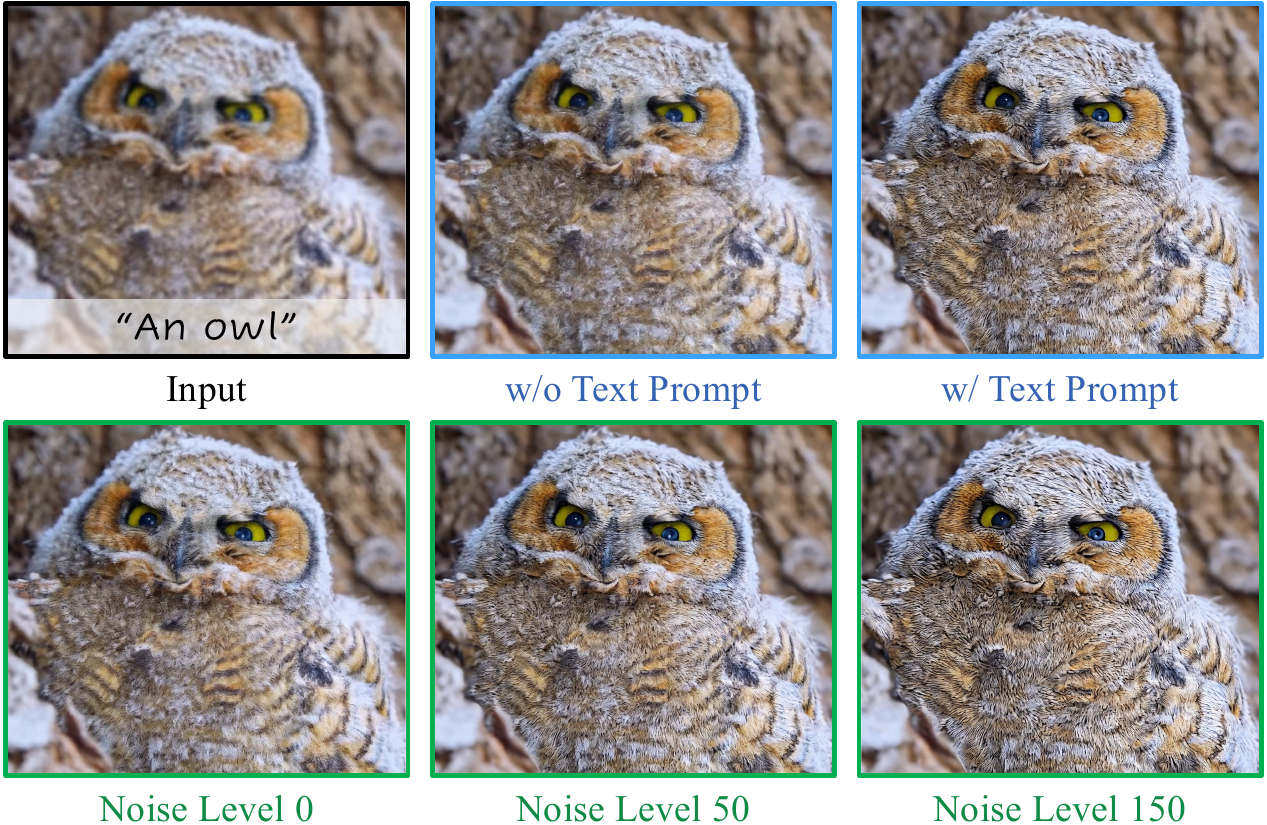}
    \vspace{-3mm}
    \caption{
    Comparison of various conditions of text prompts and different levels of noise. \textbf{(Zoom-in for best view)}
    }
    \vspace{-7mm}
\label{fig:conditions}
\end{center}
\end{figure}
Upscale-A-Video is trained on video data with either labeled prompts or null prompts and thus can handle both scenarios.
We examine the use of classifier-free guidance~\cite{ho2022classifier} to improve the visual quality during sampling.
As shown in Figure~\ref{fig:conditions}, compared with the null prompt, adopting proper text prompts can significantly enhance the perceptual quality with more faithful details.

\noindent {\bf Noise Level.} 
It is observed that the level of noise added to the input can affect the performance of our approach.
As shown in Fig.~\ref{fig:conditions}, with low noise level, the generated results tend to be suboptimal with blurry details.
However, a noise level that is too large may result in oversharpening.

%% file: 5_conclusion.tex

\vspace{0mm}
\section{Conclusion}
\label{sec:conclusion}
While diffusion models have achieved impressive performance on a wide range of image tasks, their applications on video tasks, particularly real-world VSR remain challenging and understudied. 
In this paper, we present Upscale-A-Video, a novel approach to exploit image diffusion prior for real-world VSR while avoiding temporal discontinuities drawn from the inherent randomness during sampling process. 
Specifically, we enhance the temporal coherence by proposing a novel local-global temporal strategy within the latent diffusion framework.
We additionally devote our efforts to achieving a trade-off between fidelity and quality by enabling texture creation with text prompts and noise level control, further facilitating the practical use in real-world scenarios.
We believe that our exploration would lay a good foundation for future works.


%% file: X_suppl.tex
\clearpage
\renewcommand\thesection{\Alph{section}}
\onecolumn
\setcounter{section}{0}


%
\begin{center}
	\Large\textbf{{Appendix}}\\
	\vspace{8mm}
\end{center}

{   
    \hypersetup{linkcolor=blue}
    \tableofcontents
}

\clearpage
\section{Architecture}
\label{sec:arch}
\subsection{Hyperparameters of Network}
Following the hyperparameter table style from Latent Diffusion~\cite{rombach2022high}, Table~\ref{tab:arch} provides an overview of the hyperparameters of the pretrained SD $\times$4 Upscaler~\cite{sdupscaler} and our inserted temporal layers. Our codes and models will be publicly released.

%

\begin{table}[h]
\caption{Hyperparameters for the pretrained SD $\times$4 Upscaler~\cite{sdupscaler} (including U-Net and VAE) and our inserted temporal layers. We train our model on the video patches of size $320 \times 320$ with 8 frames.}
\centering
\begin{minipage}[t]{0.33\textwidth}
\centering
\renewcommand{\arraystretch}{1}
\renewcommand{\tabcolsep}{2.05mm}
\resizebox{\linewidth}{!} {
\begin{tabular}{lc}
\toprule
Hyperparameter &  \textit{U-Net} \\ 
\midrule
Training patch shape   &  $8\times 320\times 320 \times 3$ \\
$f$  & 4 \\
$z$-shape   &  $8\times 80\times 80 \times 4$ \\
Channels & 256 \\
Depth & 2 \\
Channel multiplier & 1, 2, 2, 4 \\
Attention resolutions & 40, 20, 10 \\
Head number & 8 \\
\midrule
Embedding dimension & 1024 \\
CA resolutions & 40, 20, 10 \\
CA sequence length & 77 \\
\bottomrule
\end{tabular}
}
\label{tab:arch}
\end{minipage}
\hspace{8mm}
\begin{minipage}[t]{0.37\textwidth}
\centering
\renewcommand{\arraystretch}{1}
\renewcommand{\tabcolsep}{2.05mm}
\resizebox{\linewidth}{!} {
\begin{tabular}{lc}
\toprule
Hyperparameter &  \textit{VAE} \\
\midrule
$f$  & 4 \\
Channels & 128 \\
Channel multiplier & 1, 2, 4 \\
\bottomrule
\vspace{4.5mm}\\
\toprule
Hyperparameter &  \textit{Temporal Layers} \\
\midrule
Temporal Attention resolutions & 40, 20, 10 \\
Head number & 8 \\
Positional encoding & RoPE~\cite{llama} \\
\midrule
3D CNN kernel size & 3, 1, 1 \\
\bottomrule
\end{tabular}
}
\label{tab:arch}
\end{minipage}

\end{table}

\section{Dataset}
\subsection{YouHQ Dataset}
\begin{figure*}[h]
\begin{center}
    \includegraphics[width=.55\linewidth]{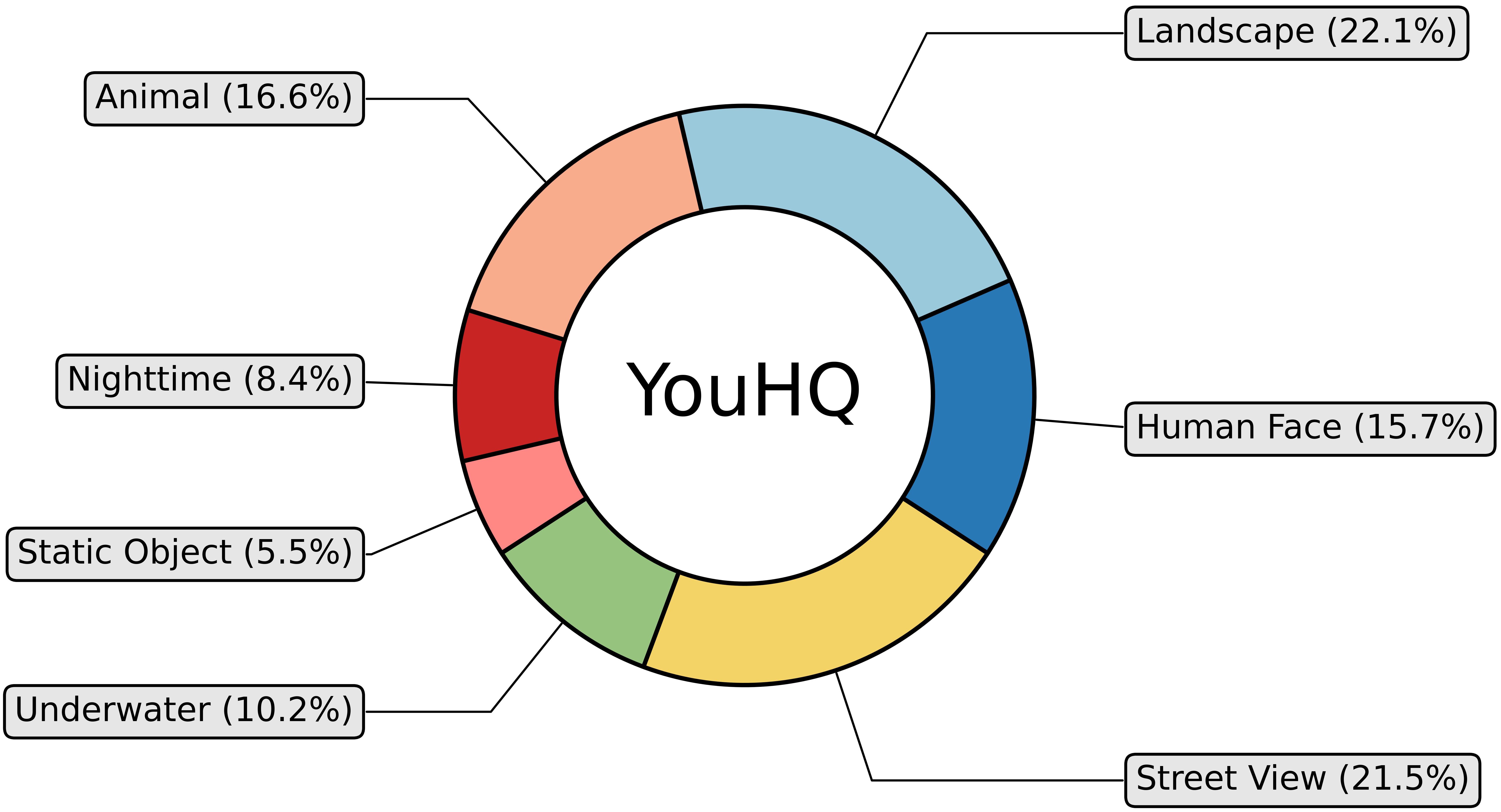}
    \vspace{-1mm}
    \caption{
        YouHQ Dataset Distribution. It consists of around 37,000 video clips with a diverse category of scenarios, including street view, landscape, animal, human face, static object, underwater, and nighttime scene. 
    }
    \vspace{-2mm}
\label{fig:youhq}
\end{center}
\end{figure*}

In order to enhance the training of our VSR model using higher-quality videos, we collect a large-scale high-definition ($1080\times 1920$) dataset from YouTube, consisting of around 37,000 video clips. The YouHQ dataset encompasses a diverse category of scenarios, including street view, landscape, animal, human face, static object, underwater, and nighttime scene. Fig.~\ref{fig:youhq} illustrates the distribution of this dataset. 

\section{More Details on Training and Inference}
\subsection{Training Strategy for Watermark Removal}
\begin{figure*}[h]
\begin{center}
    \vspace{-2mm}
    \includegraphics[width=.99\linewidth]{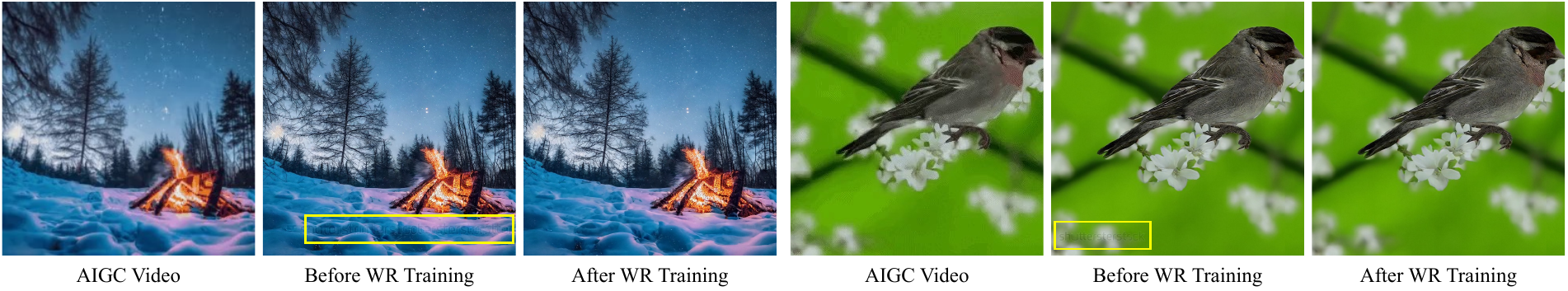}
    \vspace{-1mm}
    \caption{
        Comparison before and after watermark removal (WR) training. In the second stage, watermark removal training is performed only on the YouHQ dataset, effectively removing the watermark introduced by the first-stage training (indicated by the yellow boxes).
    }
    \vspace{-2mm}
\label{fig:remove_watermark}
\end{center}
\end{figure*}

We divide the training of the U-Net model into two phases. In the first stage, we train the U-Net using both the WebVid10M~\cite{webvid} and our introduced YouHQ datasets for 70k iterations. In the second stage, to eliminate the impact of watermarks in the WebVid10M data on the results, we conduct an additional 10k iterations of training using only the YouHQ dataset. Fig.~\ref{fig:remove_watermark} showcases the comparisons before and after the second training phase for watermark removal. 
\subsection{Inference at Arbitrary Resolution and Length}
Our model can perform inference on videos of arbitrary scales and lengths. 
This is achieved by training our model in a patch-wise manner and using the input video as a strong condition. As a result, our model effectively retains its inherent convolutional characteristics. 
Therefore, it does not impose strict input resolution requirements. Considering memory constraints, we crop the input video into multiple overlapping patches, process them separately, and finally combine the enhanced patches together.
Regarding the temporal dimension, at each diffusion step, we cut the video into clips with overlapping frames for inference. The latent features from these overlapping frames are averaged and then passed to the next diffusion step.
\subsection{Color Correction}

As noted in previous studies~\cite{choi2022perception, wang2023exploiting}, diffusion models are prone to experiencing color shift artifacts. To address this issue, we finetune the VAE-Decoder using the input as a condition, which can help maintain consistency in low-frequency information, such as color. Additionally, we have observed that incorporating a training-free \textit{wavelet color correction} module~\cite{wang2023exploiting} can further enhance color consistency in the results. As shown in Table~\ref{tab:sd_prior}, when applying wavelet color correction, our method yields slightly higher fidelity results, as indicated by improved PSNR, SSIM, and IPIPS scores.

\section{More Results}
\label{sec:result}
\subsection{User Study}

\begin{figure*}[h]
\begin{center}
    \includegraphics[width=.55\linewidth]{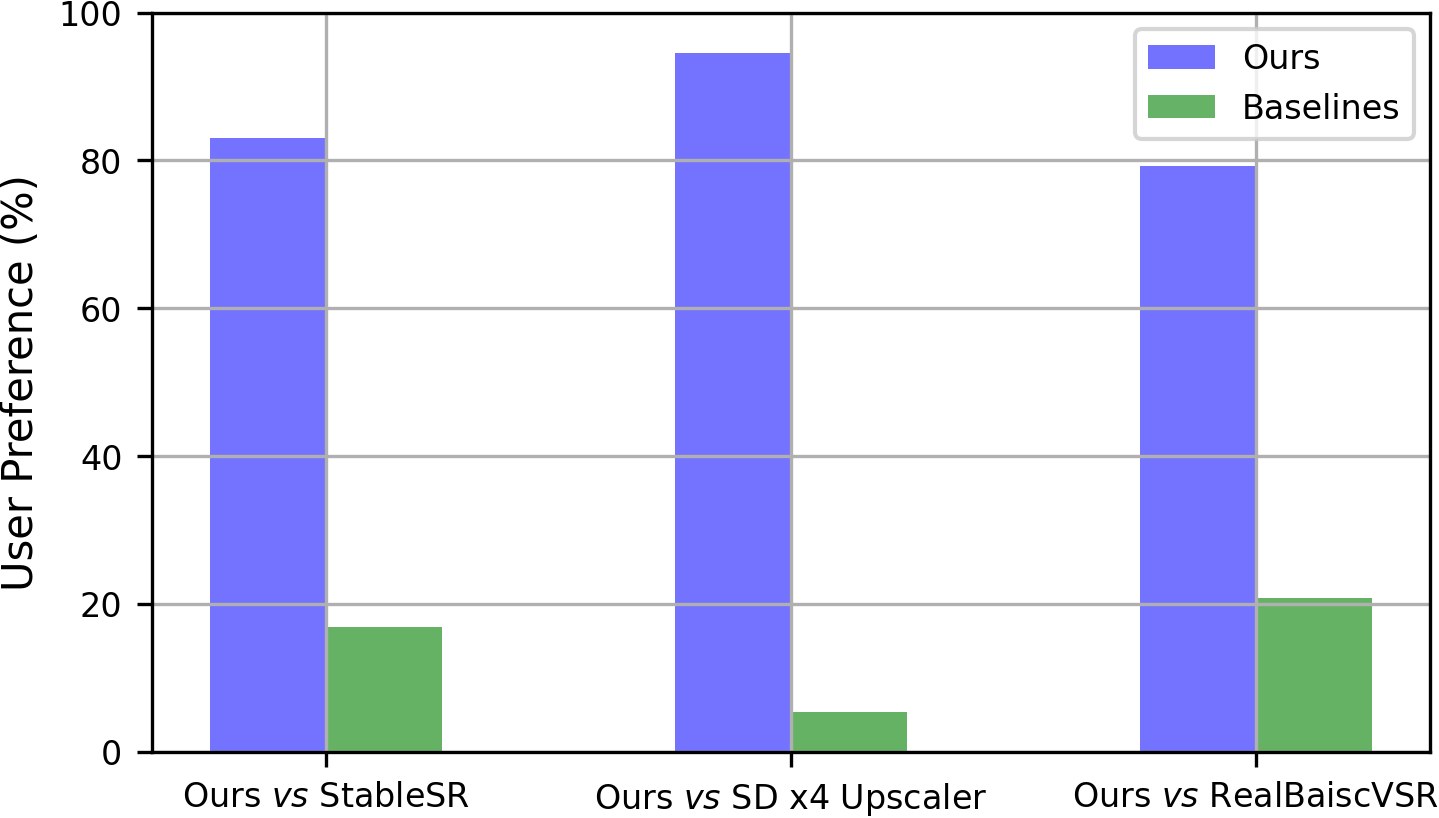}
    \vspace{-1mm}
    \caption{
        User study results. Our Upscale-A-Video is preferred by human voters over other methods.
    }
    \vspace{-6mm}
\label{fig:user_study}
\end{center}
\end{figure*}

For further comprehensive comparisons, we carried out a user study that evaluated the results of both real-world and AIGC videos. We included four different methods in this study, consisting of two diffusion-based image super-resolution methods, \ie, StableSR~\cite{wang2023exploiting} and SD $\times$4 Upscaler~\cite{sdupscaler}, along with a CNN-based video super-resolution method, \ie, RealBasicVSR~\cite{chan2022investigating}. We invite a total of 20 participants for this user study. Each volunteer was presented with a set of 10 randomly selected video triplets, which included an input video, the result obtained from one of the compared methods, and our result. Their task was to choose the visually superior enhanced video from the given options. The user study findings, depicted in Fig.~\ref{fig:user_study}, reveal a clear preference among the volunteers for our results over those produced by other methods.

\clearpage
\subsection{Ablation on Different Pretrained Priors}
\begin{figure*}[h]
\begin{center}
    \includegraphics[width=.95\linewidth]{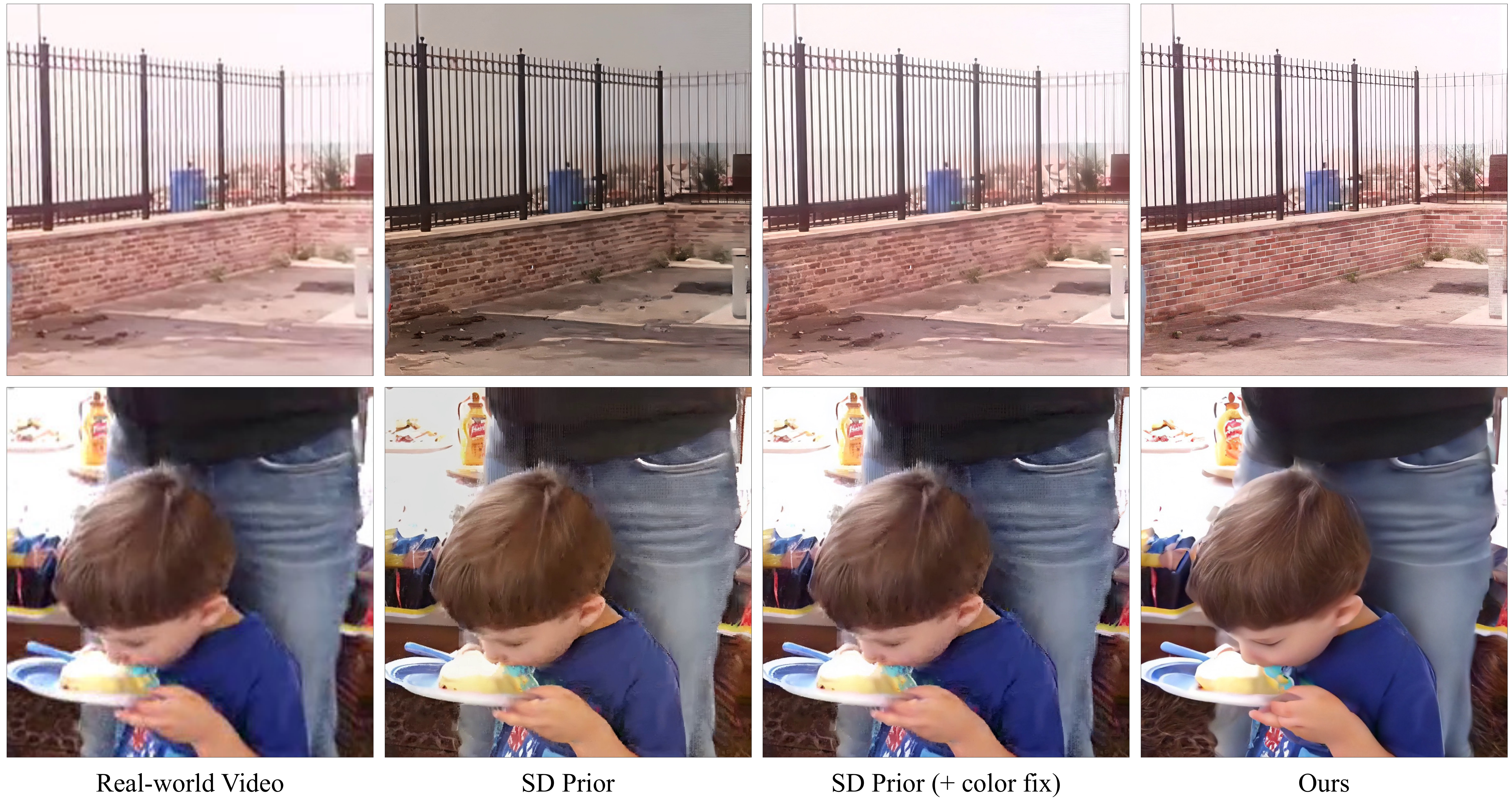}
    \vspace{-2mm}
    \caption{
        Visual comparison on variant networks with different pretrained priors (\ie, Stable Diffusion (SD)~\cite{rombach2022high} and SD $\times$4 Upscaler~\cite{sdupscaler}). The variant network based on SD often suffers from color shift, requiring additional color correction, and may also lead to unexpected artifacts, such as the child's face in the second example. Furthermore, Our model shows superior generative capabilities compared to the SD-based baseline, \eg, in the first example, our model successfully restores the wall, whereas the SD-based model fails to do so.
    }
    \vspace{-2mm}
\label{fig:sd_prior}
\end{center}
\end{figure*}

Recent studies~\cite{wang2023exploiting, lin2023diffbir} have shown that the large text-to-image Stable Diffusion (SD)~\cite{rombach2022high} is highly effective as a generative prior for blind image restoration tasks. In contrast to these works, we choose to employ a pretrained text-guided image upscaling model, \ie, SD $\times$4 Upscaler~\cite{sdupscaler}, as our prior for the video super-resolution (VSR) task. We have also employed Stable Diffusion (SD) as the prior for retraining the network and compared the results of these two different priors for the VSR task. As indicated in Table~\ref{tab:sd_prior}, our model based on the SD $\times$4 Upscaler demonstrates clear advantages in terms of restoration fidelity (PSNR, SSIM, and LPIPS) and temporal consistency ($E^*_{warp}$). It is important to note that the variant network based on SD exhibits a more noticeable color shift issue after training, which necessitates the use of the 'wavelet color correction' module~\cite{wang2023exploiting} for correction. However, even with this correction, our model outperforms the variant using SD as the prior. It is worth mentioning that when applying wavelet color correction, our model also achieves higher fidelity results in terms of PSNR, SSIM, and LPIPS. Additionally, Fig.~\ref{fig:sd_prior} provides visual comparison results for better illustration.

\begin{table}[h]
\caption{Ablation study of different pretrained priors, \ie, Stable Diffusion~\cite{rombach2022high} and SD $\times$4 Upscaler~\cite{sdupscaler}, on YouHQ40 test set. Our Upscale-A-Video based on the SD $\times$4 Upscaler showcases clear advantages in terms of restoration fidelity (PSNR, SSIM, and LPIPS) as well as temporal consistency ($E^*_{warp}$).}
\centering
\vspace{-2mm}
\renewcommand{\arraystretch}{1.15}
\renewcommand{\tabcolsep}{2.05mm}
\resizebox{0.8\linewidth}{!} {
\begin{tabular}{c|cc|cc}
\toprule
Metrics & Stable Diffusion & Stable Diffusion (+ color fix) &  \textbf{SD $\times$4 Upscaler}  & \textbf{SD $\times$4 Upscaler (+ color fix)}\\ 
\midrule
 PSNR $\uparrow$   &  19.03  & 23.81 & 25.83 & 26.07 \\
 SSIM $\uparrow$  & 0.590  & 0.632  & 0.733   &  0.737 \\ 
 LPIPS $\downarrow$  & 0.383  & 0.343  & 0.268   &  0.267 \\ 
 \hline
 $E^*_{warp} \downarrow$ & 1.821 & 1.707  &  0.737   &  0.738  \\ \bottomrule
\end{tabular}
}
\label{tab:sd_prior}
\end{table}

\clearpage
\subsection{Ablation on Positions of Recurrent Latent Propagation Module}
\begin{figure*}[h]
\begin{center}
    \includegraphics[width=.99\linewidth]{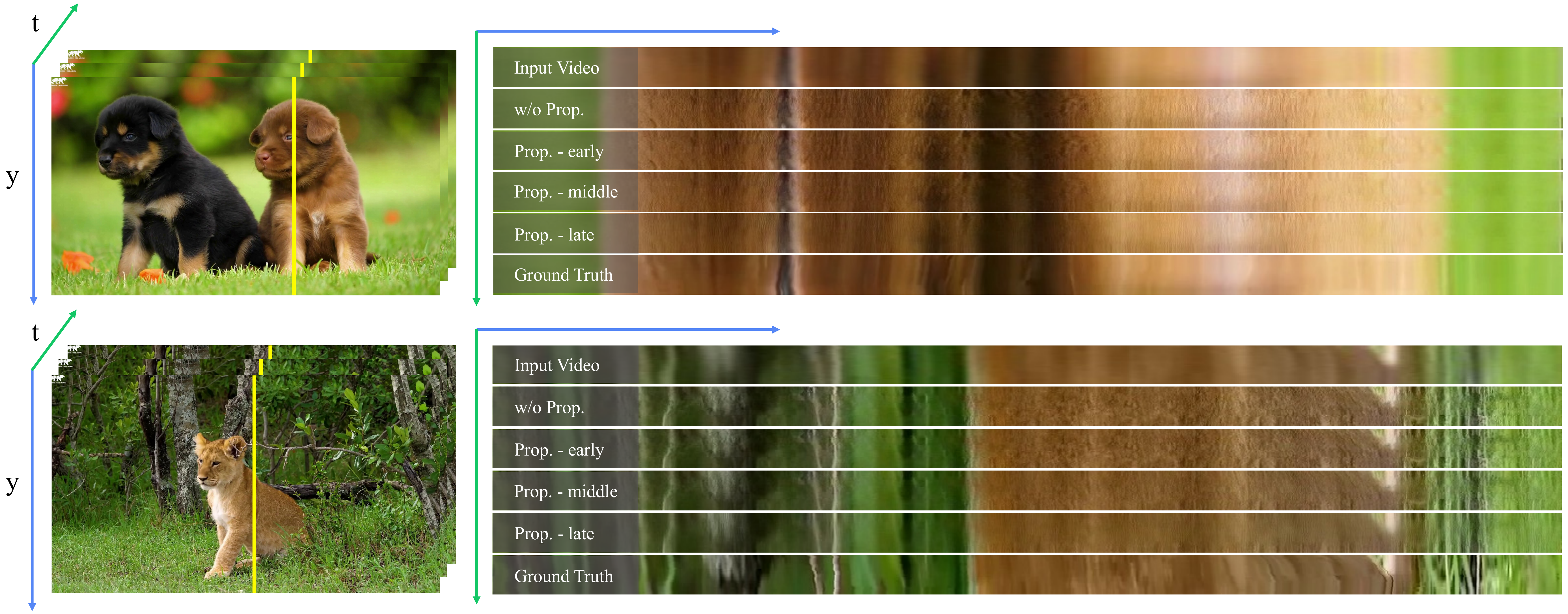}
    \vspace{-2mm}
    \caption{
        Visual comparison on temporal profile with different positions of recurrent latent propagation module.
    }
    \vspace{-2mm}
\label{fig:prop_position}
\end{center}
\end{figure*}

As discussed in Sec.~{\textcolor{red}{3.3}} of the main manuscript, it is not necessary to employ the recurrent latent propagation module during every diffusion step in the inference process. Instead, we have the flexibility to choose specific steps for latent propagation and aggregation. Here, we showcase the performance variations when placing this module at different positions, evaluating on the YouHQ40 test set. The results presented in Table~\ref{tab:prop_position} indicate that when propagation happens later in the diffusion denoising steps during inference, the warping loss tends to decrease, suggesting better temporal consistency. However, the restoration fidelity also decreases. To balance these factors, we by default choose the middle position for this propagation module. Additionally, Fig.~\ref{fig:prop_position} provides the visual comparisons of the temporal profile, illustrating that as propagation occurs later, the videos exhibit improved temporal coherence.
\begin{table}[h]
\caption{Ablation study of different positions of recurrent latent propagation module on the YouHQ40 test set.}
\centering
\vspace{-2mm}
\renewcommand{\arraystretch}{1.15}
\renewcommand{\tabcolsep}{2.05mm}
\resizebox{0.56\linewidth}{!} {
\begin{tabular}{c|cccc}
\toprule
\multirow{2}{*}{Metrics} & w/o prop. & early prop. &  middle prop.  & late prop. \\ 
       & - & \{4, 5, 6, 7\} &  \{14, 15, 16, 17\}  & \{24, 25, 26, 27\} \\ 
\midrule
 PSNR $\uparrow$   &  23.82  & 24.18 & 24.53 & 24.10 \\
 SSIM $\uparrow$  & 0.639  & 0.646  & 0.671   &  0.670 \\ 
 \hline
 $E^*_{warp} \downarrow$ & 2.398 & 1.931  &  0.638   &  0.618  \\ \bottomrule
\end{tabular}
}
\label{tab:prop_position}
\end{table}
\subsection{Effectiveness of Text Prompt}
\begin{figure*}[t]
\begin{center}
    \includegraphics[width=.95\linewidth]{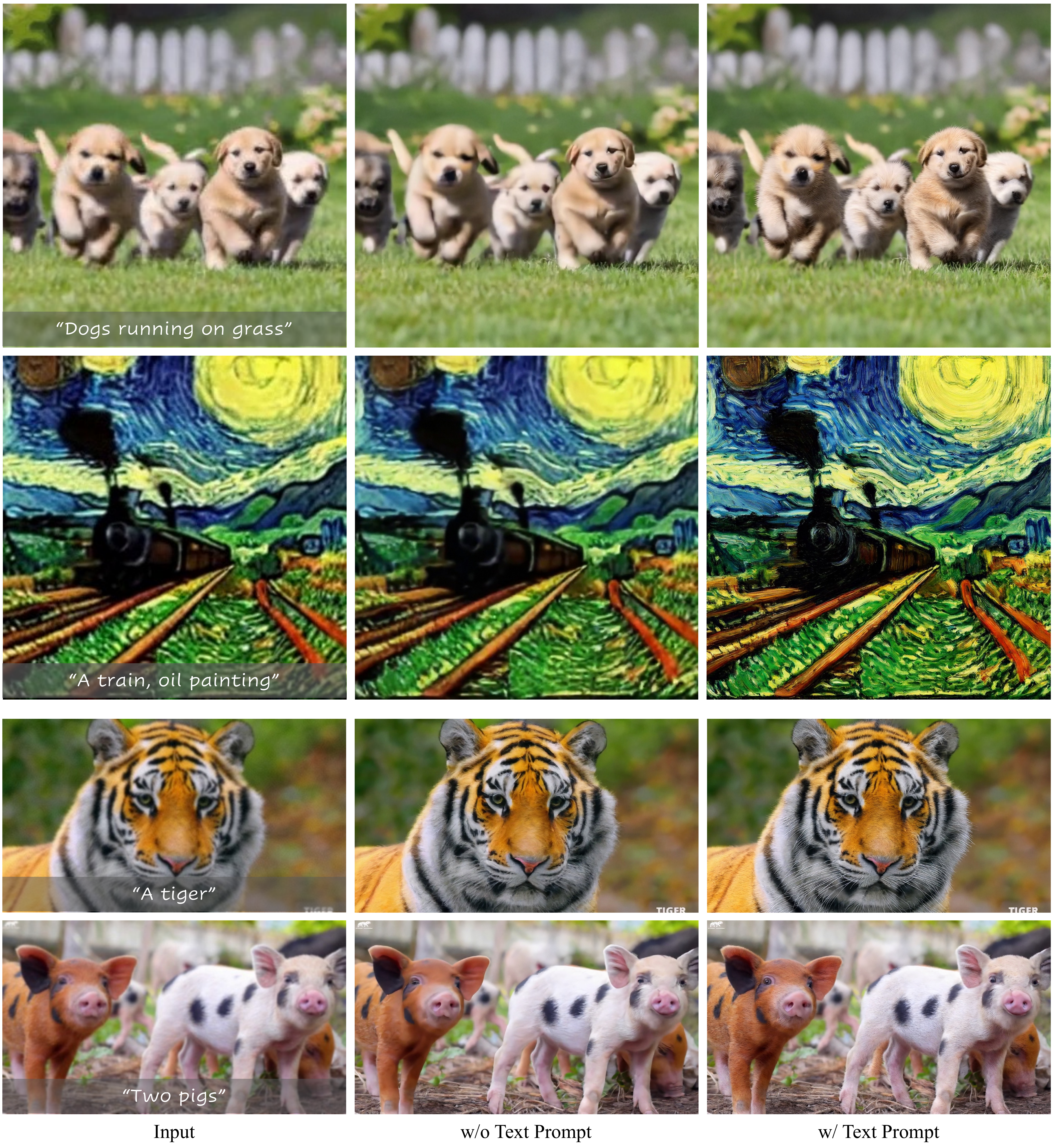}
    \vspace{-2mm}
    \caption{
        Visual comparison of using proper text prompts and empty prompts. When employing the classifier-free guidance~\cite{ho2022classifier}, using proper text prompts as guidance can significantly improve the visual quality and realism, resulting in finer details. This improvement is observed in both real-world scene videos (the last two rows) and AIGC videos (the first two rows).
    }
    \vspace{-2mm}
\label{fig:text_prompt}
\end{center}
\end{figure*}

Upscale-A-Video is trained using video data that includes labeled prompts or no prompts, allowing it to work effectively in both situations.
However, when employing the classifier-free guidance approach~\cite{ho2022classifier}, utilizing proper text prompts as guidance can noticeably enhance the visual quality.
As illustrated in Fig.~\ref{fig:text_prompt}, the use of appropriate text prompts leads to significantly improved results with finer and more faithful details compared to using empty prompts.

%

%
\clearpage
\subsection{More Qualitative Comparisons}
\begin{figure*}[h]
\begin{center}
    \includegraphics[width=0.99\linewidth]{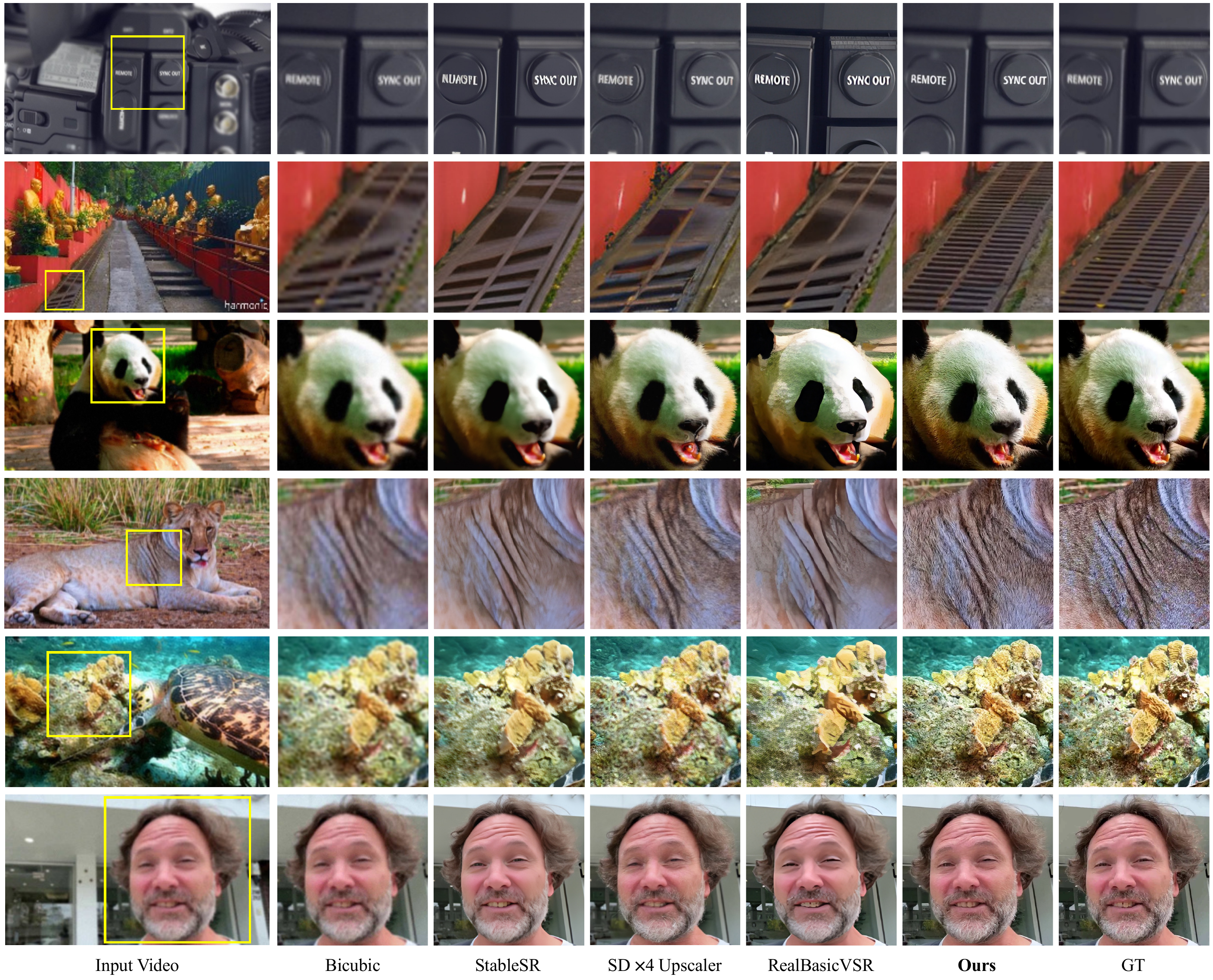}
    \vspace{-1mm}
    \caption{
        Qualitative comparisons on synthetic datasets. Our Upscale-A-Video exhibits promising enhanced results with more details and heightened realism. (\textbf{Zoom in for best view.})
    }
    \vspace{-2mm}
\label{fig:results_synthetic}
\end{center}
\end{figure*}

\begin{figure*}[t]
\begin{center}
    \includegraphics[width=0.99\linewidth]{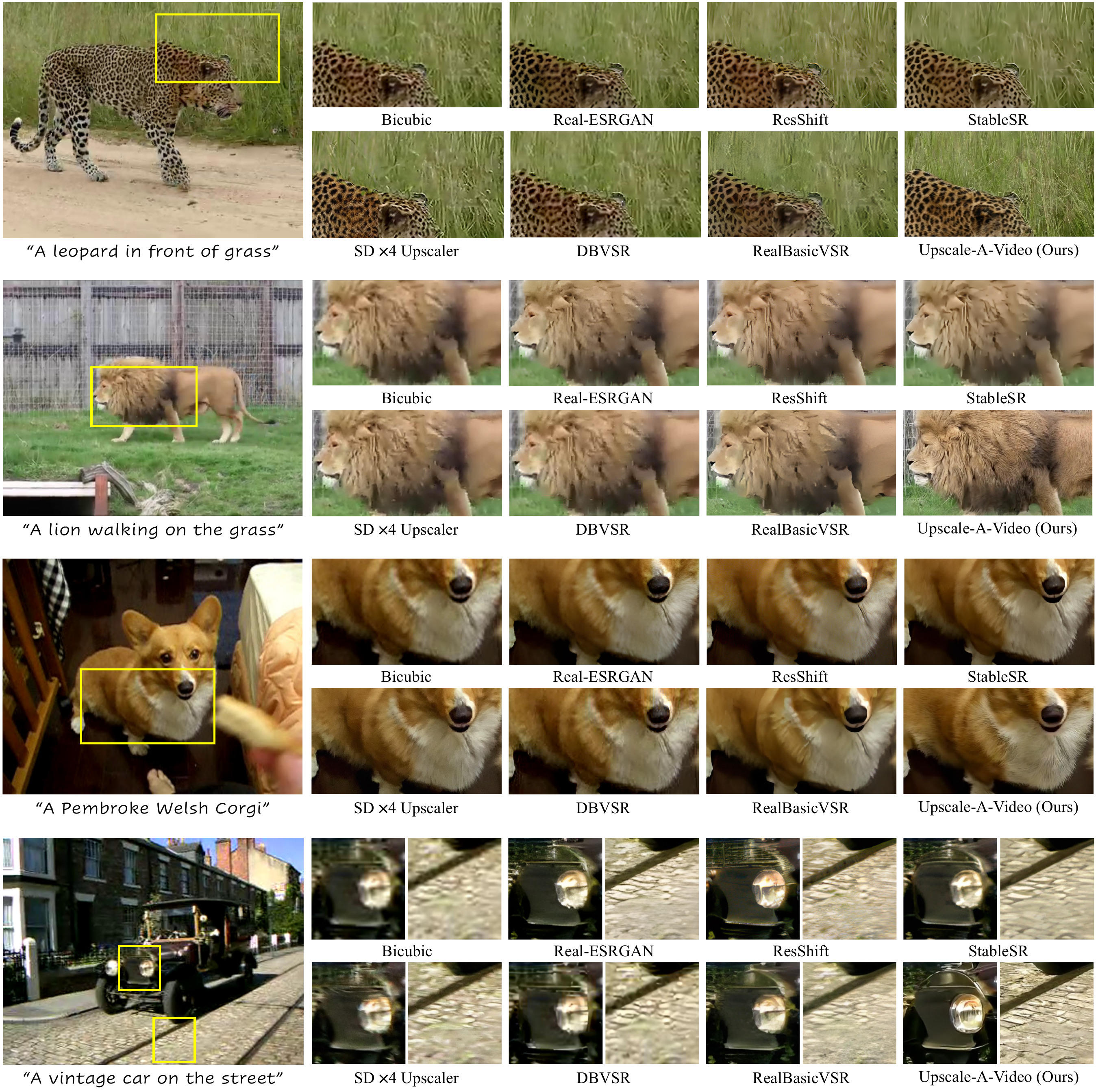}
    \vspace{-1mm}
    \caption{
        Qualitative comparisons on real-world videos. Our Upscale-A-Video produces promising improvements, delivering increased detail and heightened realism. (\textbf{Zoom in for best view.})
    }
    \vspace{-2mm}
\label{fig:results_real}
\end{center}
\end{figure*}

\begin{figure*}[t]
\begin{center}
    \includegraphics[width=0.99\linewidth]{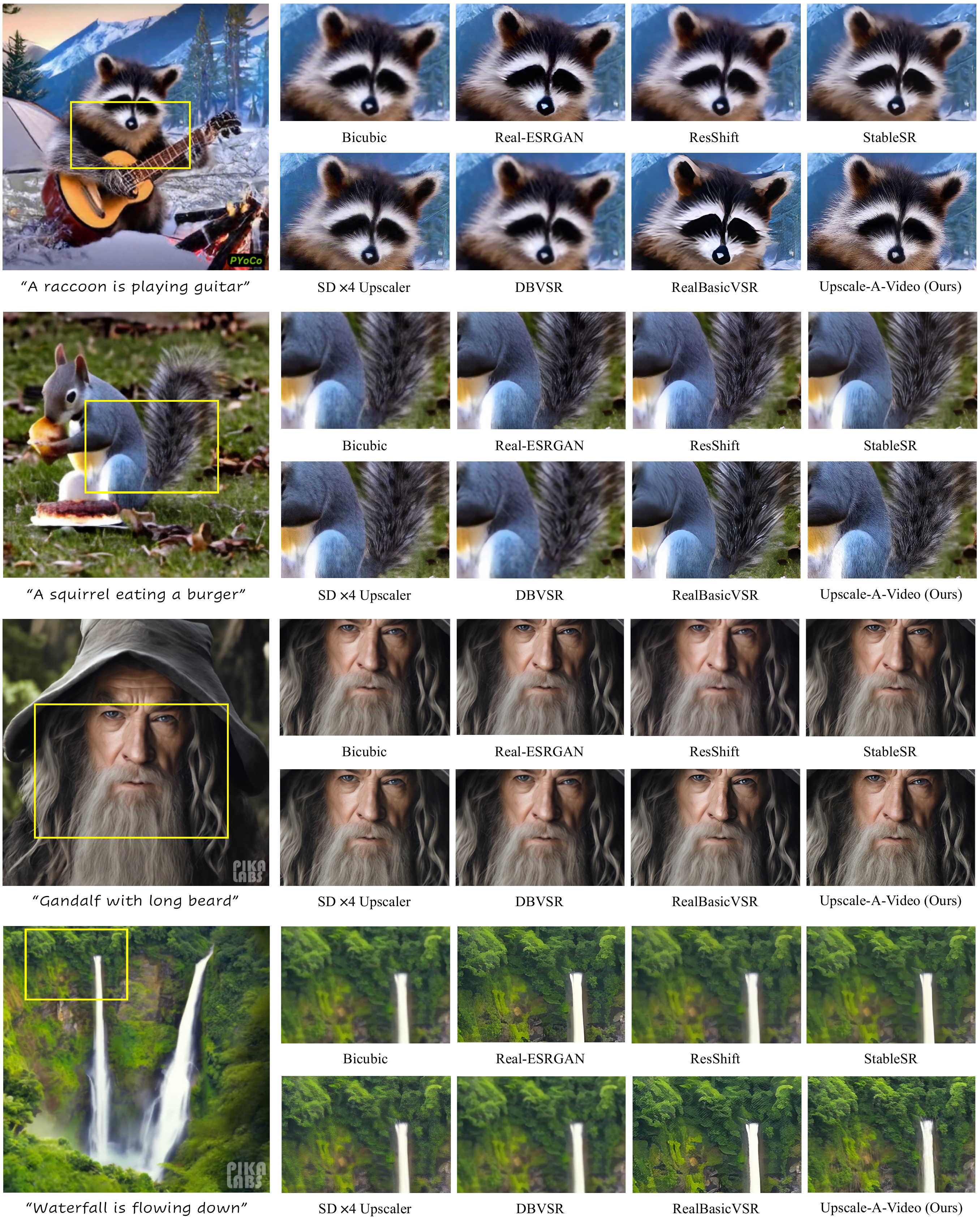}
    \vspace{-1mm}
    \caption{
        Qualitative comparisons on AIGC videos. When guided by input text prompts, our Upscale-A-Video exhibits promising video results with more details and enhanced realism. (\textbf{Zoom in for best view.})
    }
    \vspace{-2mm}
\label{fig:results_aigc}
\end{center}
\end{figure*}

In this section, we provide additional visual comparisons of our method with the state-of-the-art methods, including Real-ESRGAN~\cite{wang2021realesrgan}, SD $\times$4 Upscaler~\cite{sdupscaler}, ResShift~\cite{yue2023resshift}, StableSR~\cite{wang2023exploiting}, DBVSR~\cite{pan2021deep}, and RealBasicVSR~\cite{chan2022investigating}.
Fig.~\ref{fig:results_synthetic}, Fig.~\ref{fig:results_real}, and Fig.~\ref{fig:results_aigc} present the visual results on synthetic, real-world, and AIGC videos, respectively.


\subsection{Video Demo}
We also offer a demo video \href{https://youtu.be/b9J3lqiKnLM?si=lM93wkr7VS3m_57b}{[{\color{red}{Upscale-A-Video-demo.mp4}}]} to showcase more video results and comparisons, which are evaluated on synthetic, real-world, and AIGC videos.